# Decision Lists for English and Basque

David Martinez
IXA NLP Group
University of the Basque Country
649 pk. 20.080
Donostia. Spain.
jibmaird@si.ehu.es

Eneko Agirre
IXA NLP Group
University of the Basque Country
649 pk. 20.080
Donostia. Spain.
eneko@si.ehu.es

**Abstract**

In this paper we describe the systems we developed for the English (lexical and all-words) and Basque tasks. They were all supervised systems based on Yarowsky's Decision Lists. We used Semcor for training in the English all-words task. We defined different feature sets for each language. For Basque, in order to extract all the information from the text, we defined features that have not been used before in the literature, using a morphological analyzer. We also implemented systems that selected automatically good features and were able to obtain a prefixed precision (85%) at the cost of coverage. The systems that used all the features were identified as BCU-ehu-dlist-all and the systems that selected some features as BCU-ehu-dlist-best.

## 1   Introduction

Our group took part in three tasks in Senseval-2001: all-words and lexical sample for English, and lexical sample for Basque. We applied the same algorithm in all tasks, but using different feature sets. The method we used was based on Yarowsky's Decision Lists (Yarowsky, 1994).

We have to mention that different motivations were pursued when working for English or for Basque. In the last years, our work for English has focused on studying the contribution of different kinds of features to WSD (Agirre and Martinez, 2000; Agirre and Martinez, 2001a) and on analyzing different knowledge types on a common setting (Agirre and Martinez, 2001b).

The English tasks gave us the chance to compare the performance of our method with state-of-the-art systems. Unfortunately, due to time constrains, we could not train the system with syntactic and semantic features, as was our goal. The systems we used for the English tasks were trained with topical and local features already mentioned in the literature (Yarowsky, 1994).

In the English lexical sample, we presented two systems: one trained with all the features (BCU-ehu-dlist-all) and another that selected automatically good features and was able to obtain a prefixed precision (85%) at the cost of coverage (BCU-ehu-dlist-best). In the all-words exercise, we presented only one system, which used all the features and was trained using Semcor (previously we mapped automatically WN 1.6 senses with the WN 1.7 Pre-release). The features and the systems will be described in Section 3.

The Basque task presented interesting challenges for us. Previous work in WSD was performed using MRDs, but this was our first approach to the problem using Decision Lists. The Basque language has some particularities that make the selection of features a difficult task. First of all, Basque is an agglutinative language, and some syntactic information is given by inflectional suffixes. Therefore, it is necessary a powerful morphologic analysis of the text in order to identify the lemma and the different parts of each word. Also, phenomena like noun ellipsis have to be taken into account.

We had the tools to perform a deep morphological analysis of the text (Urkia, 1997) and we were able to define a richer feature set than the one used for English. The complex structure of the analysis allowed constructing different feature types, which should be studied in detail. However, our approach was to define many features and integrate them together in the system, expecting that the Decision List algorithm would be powerful enough to choose the best ones in each case. We will explain the feature set in Section 4. We also presented two systems for Basque: one using all the features and another selecting good features (those above a threshold of 85% precision).

Following this introduction, we will briefly explain the Decision List algorithm in Section 2. Section 3 will be devoted to the English lexical task and Section 4 to the English all-words task. The Basque lexical task will be described in Section 5. The results obtained by our systems will be discussed in Section 6. Finally, we will resume our conclusions in Section 7.

## 2 Decision Lists

Decision lists as defined in (Yarowsky, 1994) are simple means to resolve ambiguity problems. With the addition of some hierarchical structure, it was one of the most successful systems on the Senseval-1 exercise. The training data is processed to extract the features, which are weighted with a log-likelihood measure. The list of all features ordered by the log-likelihood values constitutes the decision list. We adapted the original formula in order to accommodate ambiguities higher than two.

$$weight(sense_i, feature_k) = Log(\frac{\Pr(sense_i | feature_k)}{\sum_{j \neq i} \Pr(sense_j | feature_k)})$$

It is not clear what to do when all weights of the senses for the given feature are below 0. We decided to delete such features from the decision lists.

When testing, the decision list is checked in order and the feature with highest weight that is present in the test sentence selects the winning word sense. The probabilities have been estimated using the maximum likelihood estimate, smoothed using a simple method: when the denominator in the formula is 0 we replace it with 0.1. The estimates can be improved using more sophisticated smoothing techniques.

## 3 English lexical-sample task

In the English lexical sample task, we presented two systems: BCU-ehu-dlist-all and BCU-ehu-dlist-best.

### 3.1. BCU-dlist-ehu-all

We trained our Decision List algorithm using local and global features:

- Local features: bigrams and trigrams around the target word, consisting on lemmas or word forms or parts of speech. Also a bag of lemmas constructed using the content words in a ±4 word window around the target.
- Global features: a bag of lemmas with the content words included in the whole context provided for the target word.

We did not use the tags P and U. There was no special treatment for multiword detection.

### 3.2. BCU-dlist-ehu-best

In this case, instead of using the whole set of features, only the best features for each word were chosen. Features that had a precision above a threshold of 85% were automatically selected running the system on the training data, using 10 fold cross validation. With this second system we wanted to guarantee high precision for each word, at the cost of coverage.

## 4 English all-words task (BCU-dlist-ehu-all)

In the all-words exercise, we presented only one system, which used all the features and was trained using Semcor. A mapping between the senses in Semcor (tagged with WordNet 1.6 senses) and WordNet 1.7 was performed automatically for nouns and verbs; only words with those parts-of-speech were treated (we could not finish the mapping for adjectives on time).

## 5 Basque lexical-sample task (BCU-dlist-ehu-all and BCU-dlist-ehu-best)

As mentioned before, Basque is an agglutinative language, and syntactic information is given by inflectional suffixes. The morphological analysis of the text is a necessary previous step in order to select informative features. In Basque, the determiner, the number and the declension case are appended to the last element of the phrase. In order to include this information in our representation, we have to use more rich features than those defined for English. When defining our feature set for Basque, we tried to introduce the same knowledge that is represented by features that work well for English.

We will describe our feature set with an example: for the phrase "**elizaren arduradunei**" (which means "to the directors of the church") we get the following analysis:

| eliza  | ren    | arduradun | ei            |
|--------|--------|-----------|---------------|
| church | of the | director  | to the +plural|

The order of the words is inverse in English. We extract the following information for each word:

elizaren:
    Lemma: eliza (church)
    PoS: noun
    Declension Case: genitive (of)
    Number: singular
    Determiner mark: yes
arduradunei:
    Lemma: arduradun (director)
    PoS: noun
    Declension Case: dative (to)
    Number: plural
    Determiner mark: yes

We will assume that *eliza* (*church*) is the target word. Words and lemmas are shown in lowercase and the other information in uppercase. As local features we defined different types of unigrams, bigrams, trigrams and a window of ±4 words. The unigrams were constructed combining word forms, lemmas, case, number, and determiner mark. We defined 4 kinds of unigrams:

Uni_wf0 elizaren
Uni_wf1 eliza   SING+DET
Uni_wf2 eliza   GENITIVE
Uni_wf3 eliza   SING+DET   GENITIVE

As for English, we defined bigrams based on word forms, lemmas and parts-of-speech. But in order to simulate the bigrams and trigrams used for English, we defined different kinds of features. For word forms, we distinguished two cases: using the text string (Big_wf0), or using the tags from the analysis (Big_wf1). The word form bigrams for the collocation "elizaren arduradunei" are shown below. In the case of the feature type "Big_wf1", the information is split in three features:

Big_wf0 elizaren arduradunei

Big_wf1 eliza GENITIVE
Big_wf1 GENITIVE arduradun_PLUR+DET
Big_wf1 arduradun_PLUR+DET DATIVE

Similarly, depending on the use of the declension case, we defined three kinds of bigrams based on lemmas:

Big_lem0 eliza arduradun

Big_lem1 eliza GENITIVE
Big_lem1 GENITIVE arduradun
Big_lem1 arduradun DATIVE

Big_lem2 eliza_GENITIVE
Big_lem2 arduradun_DATIVE

The bigrams constructed using Part-of-speech are illustrated below. We included the declension case:

Big_pos_-1 NOUN GENITIVE
Big_pos_-1 GENITIVE NOUN
Big_pos_-1 NOUN DATIVE

Trigrams are built similarly, by combining the information from three consecutive words. We also used as local features all the content words in a window of ±4 words around the target. Finally, as global features we took all the content lemmas appearing in the context, which was constituted by the target sentence and the two previous and posterior sentences.

One difficult case to model in Basque is the ellipsis. For example, the word "elizakoa" means "the one from the church". We were able to extract this information from our analyzer and we represented it in the features, using a mark as the elliptic word.

We implemented two systems; in the first one, we integrated all the features in the Decision List algorithm, expecting that the most informative ones would be chosen. The performance of the different features was not studied separately. Our second system for Basque applied feature selection in a similar way as for English.

This was our first approach to represent the Basque sentences in a feature set suitable for the Decision List algorithm. We detected some reasons that could have lowered the performance of the system:
- In Basque the word order is free. The performance of bigrams and trigrams, which have to be in fixed positions, could be affected for this fact.

- When we introduce the number, declension case, and determiner; the relation between some words that are close in the text could be lost. We have tried to overcome this by defining many features, but we did not analyze them by hand, and some could introduce noise. Deeper study of the features should be done in order to know the real performance of the method. Uncommon cases, like ellipsis, should be further examined.
- Another source of noise was the morphological analyzer, which in some cases produced very ambiguous analysis, or errors.

## 6 Results and Discussion

In the English lexical task, BCU-ehu-dlist-all scored 57.3% in precision and 98% in coverage. It beat easily the different baselines and with a simple implementation, was close in precision to more elaborate and complex systems. With BCU-ehu-dlist-best we were able to obtain a precision of 82.9% for 28% coverage. The threshold of 85% precision proved to be too high for some words, and too low for others. Besides, the chosen features had low coverage and could be applied only in a few cases.

In the English all-words task, we obtained almost the same precision as in the lexical task: 57.2%. The coverage was limited to nouns and verbs with training examples in Semcor, and reached 51% of the target words. Clearly, more training data was required to compete in recall with the best systems.

For Basque, with BCU-ehu-dlist-all we obtained 73.2% precision for 100% coverage. The system improved in almost 9 points the precision of the most frequent sense (MFS) baseline, but was two points below the best system (JHU- John Hopkins University). We have to notice that the JHU system won the lexical sample task both for Basque and for English; and while the difference in recall with our system was only 2% for Basque, it reached 8% for English. We think that the reason for this is that our feature set for Basque is better, although our ML algorithm is worse.

Finally, with BCU-ehu-dlist-best the 85% threshold worked better than for English and we reached higher coverage. We were able to obtain 84.9% precision for 57% coverage. However, again, the threshold was too high for some words (for 4 words no feature was chosen), and too low for others (easy words like "enplegu" chose the whole feature set).

The positions of our systems in the different tasks are illustrated in Table 1:

| Task | System (BCU-ehu-dlist...) | Position Precision | Position Recall |
|---|---|---|---|
| Basque lexical | best | 1st of 3 | 3rd of 3 |
| Basque lexical | all | 3rd of 3 | 2nd of 3 |
| English lexical | best | 1st of 20 | 20th of 20 |
| English lexical | all | 9th of 20 | 9th of 20 |
| English all-words | all | 7th of 21 | 14th of 21 |

**Table 1:** Classification of our systems (*version 1.5, published 28 Sep. 2001*). Fine-grained scoring. Only last versions of resubmitted systems (R) are included. Baselines are not incorporated. Only supervised systems are included in the lexical tasks.

## 7 Conclusions

In the English tasks we were able to compare a limited version of our system (with a reduced feature set) with state-of-the-art systems. We observed that with minimum work, we could obtain results above the average of the other systems. Our next goal is to test the system with semantic and syntactic features and compare the performance with other systems.

For Basque, we defined a preliminary set of features and achieved good performance. Our results were close to the best system and above the MFS baseline. In the future, we want to refine the feature set and explore other sources of information, as syntactic features.

Finally, more experiments on feature selection should be performed in order to take advantage of this technique.